\documentclass[a4paper,12pt]{article}
\usepackage{arxiv}

\usepackage[utf8]{inputenc} 
\usepackage[T1]{fontenc}    
\usepackage{hyperref}       
\usepackage{url}            
\usepackage{amsfonts}       
\usepackage{amsmath}        
\usepackage{graphicx}       
\usepackage{caption}
\usepackage{subcaption}
\usepackage{xcolor}
\usepackage{algorithm}
\usepackage{algpseudocode}
\usepackage{abstract}

\definecolor{linkclr}{cmyk}{0.973,0.957,0,0.04}

\usepackage{environ}
\newcommand{\acksection}{\section*{Acknowledgments}}
\NewEnviron{ack}{
  \acksection
  \BODY
}

\title{Hidden Knowledge: Mathematical Methods for the Extraction of the Fingerprint of Medieval Paper from Digital Images} 
\author{Tamara G. Grossmann\thanks{Department of Applied Mathematics and Theoretical Physics, University of Cambridge, Cambridge, UK \hspace*{1.8em} \texttt{tg410@cam.ac.uk}} , Carola-Bibiane Schönlieb\footnotemark[1] \hspace{1pt} and  Orietta Da Rold\thanks{Faculty of English, University of Cambridge, Cambridge, UK}}

\usepackage{url}
\usepackage{hyperref}
\hypersetup{%
        colorlinks,
        unicode=false,
        breaklinks=true,
        plainpages=false,%
        citecolor=linkclr,
        linkcolor=linkclr,
        urlcolor=linkclr}

\usepackage{listings}
\begin{document}
\maketitle

\vspace{20pt}
\begin{abstract}
Medieval paper, a handmade product, is made with a mould which leaves an indelible  imprint on the sheet of paper. This imprint includes chain lines, laid lines and watermarks which are often visible on the sheet. Extracting these features allows the identification of paper stock and gives information about chronology, localisation and movement of books and people. Most computational work for feature extraction of paper analysis has so far focused on radiography or transmitted light images. While these imaging methods provide clear visualisation for the features of interest, they are expensive and time consuming in their acquisition and not feasible for smaller institutions. However, reflected light images of medieval paper manuscripts are abundant and possibly cheaper in their acquisition. In this paper, we propose algorithms to detect and extract the laid and chain lines from reflected light images. We tackle the main drawback of reflected light images, that is, the low contrast attenuation of lines and intensity jumps due to noise and degradation, by employing the spectral total variation decomposition and develop methods for subsequent line extraction. Our results clearly demonstrate the feasibility of using reflected light images in paper analysis. This work enables the feature extraction for paper manuscripts that have otherwise not been analysed due to a lack of appropriate images. We also open the door for paper stock identification at scale.
\end{abstract}

\section{Introduction}
Paper, one of the most versatile and lasting material technologies from medieval to present days, has carried great importance throughout the ages. It has been used to record and preserve history, was the carrier of information and as such tells the story of the agents and sites that it encountered. While nowadays paper is made with large paper machines, each sheet of medieval paper was handmade by taking a sieve - the mould - and dipping it into a large vat of pulp made from fibres. When lifting the mould out of the vat, excess water drips out and the newly formed paper sheet is couched on a pile of felts, pressed and hanged to dry. The imprint of the mould left on each sheet of paper at the point of its making is like a ‘fingerprint’ which can be identified, described and traced~\cite{daRold2007}. A mould consists of horizontal (laid) wires and vertical (chain) wires in a wooden frame with additional wires twisted into a shape which produces the watermark. The thickness of the wires and distance between them, the size of the watermark and the mould vary, and this variation constitutes the signature that the mould leaves on each sheet of paper. Such fingerprint can change overtime with the use of the mould, because the wires move, however, its identification allows individual sheets of paper to be connected and grouped. Papermakers used two moulds, with often similar features, to make sheets of paper simultaneously. It has been proposed that the paper produced by these two moulds is considered a paper stock. The study of the morphological features of a paper stock has enormous potential to study chronology, localisation and movement of books and people.
\begin{figure*}[t]
    \centering
    \includegraphics[width=0.98\textwidth]{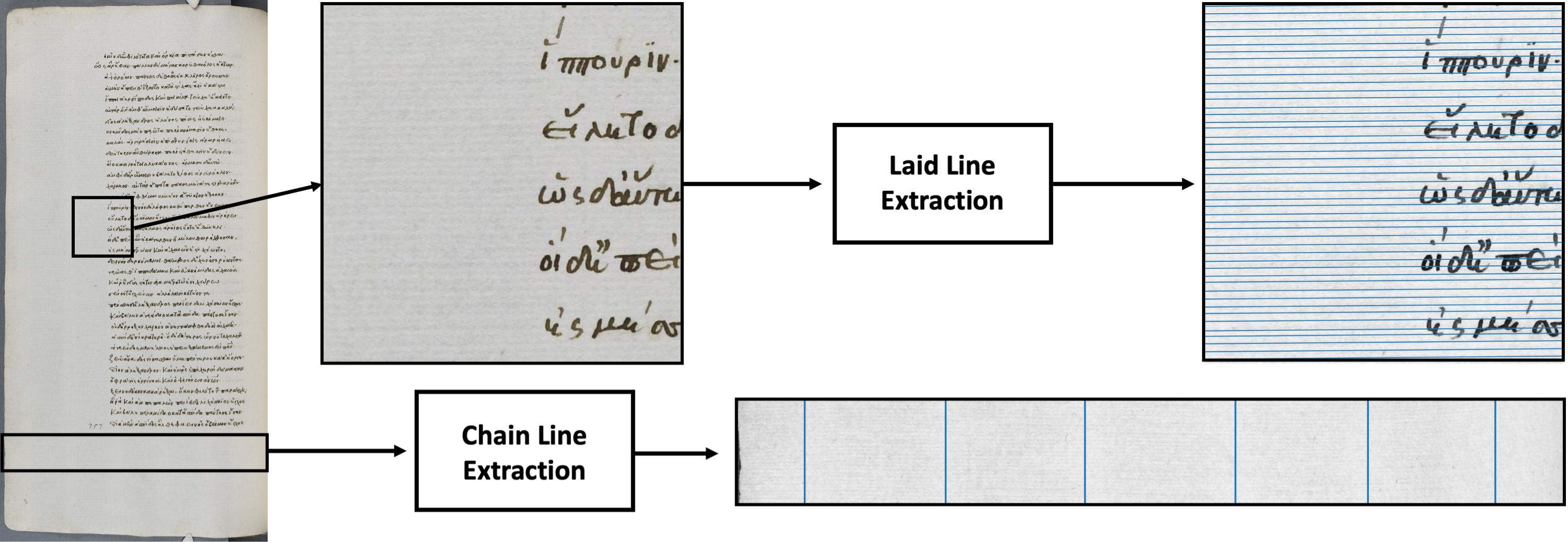}
    \caption{Overview of the features and outcomes of interest. Original image courtesy of The Parker Library, Cambridge, Corpus Christi College, MS 081: Homer, Iliad, Odyssey. Quintus of Smyrna, Posthomerica, p. 55.}
    \label{fig.abstract}
\end{figure*}

With the advent of digital technologies, a large number of paper manuscripts have been digitised in the last two decades amounting to an impressive catalogue of images ranging from radiography to transmitted and reflected light images. However, each image type comes with its own set of challenges. Radiography equipment is expensive, obtaining the images can take a significant amount of time and there are considerably less images readily available. On the other hand, transmitted light images are attainable at a lower equipment cost but they are relatively scarce and burdened with selection bias. They also contain shadows of the ink from both paper sides. Therefore, analysing transmitted light images of paper that contain written text has to include the digital removal of any ink shining through the paper, adding an extra layer of pre-processing in the image analysis pipeline. In turn, reflected light images are abundant and easily available from many archives and libraries. The analysis of reflective light images of paper is challenging due to low contrast attenuation of the expected lines, the jumps in the intensity values due to several factors (including image noise, degradation of the manuscripts, dirts and spots), and the strong penetration of the ink.

\subsection{Context}
The fingerprint of a mould is made up of at least three main paper features: the watermark, the chain lines and the laid lines. Extraction and quantification of these features has mostly been done based on radiography and transmitted light images. The work for watermark identification, measurement and extraction has been the focus of many publications. Whelan et al.~\cite{Whelan2001} proposed using morphological filters for watermark segmentation from transmitted light images. For laid paper, they additionally remove any visible laid or chain lines by Fourier filtering. However, their method is derived for paper that does not contain any writing. Hiary et al.~\cite{Hiary2006} propose to use morphological filters to separate foreground and background interference in transmitted light images, enabling the analysis of papers that have ink potentially on both sides. Afterwards, they use edge detection to extract the watermark. In contrast, Gorske et al.~\cite{Gorske2021} take a different approach in that they measure the distance between the watermark and the neighbouring chain lines instead of using the the segmented watermark for paper stock identification and matching. The Bernstein project~\cite{Rueckert2009} built a large watermark catalogue and portal that combines the digitised watermarks from many research institutions and standardised the terminology. Their platform also provides an advanced search option to filter for chain line distances around the watermark~\cite{BernsteinWebsite}. It is to note, however, that chain line distances around a watermark tend to be smaller as compared to the rest of the paper manuscript. This is due to the chain lines acting as a support to the watermark wires. The Bernstein project additionally provides software for watermark detection and extraction for radiography and transmitted light images. Watermarks as a feature of handmade paper can be classified according to iconography and this type of clustering can show differences also within the same design. However, it is well know that not all paper contains a watermark and it is important to test the extent to which chain and laid lines can also help with the identification of the mould. Therefore, focusing on chain and laid lines has its merit and it is imperative to have accurate chain and laid line extraction methods for paper stock identification. To this end, this paper focuses on chain and laid lines extraction and we will expand our methods to watermark segmentation in future work.

\subsection{State of the Art for Chain and Laid Line Extraction}
The overarching methods used for chain and laid line extraction are based on the Fourier transform and Radon transform or involve morphological filters. Van der Lubbe et al.~\cite{vanderlubbe2001} detects chain lines in radiography images using morphological filters and vertical projection, assuming the chain lines to be vertical and straight. Leveraging two paper features - chain and laid lines - van Staalduinen et al.~\cite{vanstaalduinen2006} extracts chain lines based on their shadows via Fourier transform, Radon transform and Gaussian filtering. They additionally estimate the laid line density from the Fourier frequency with the highest magnitude and then define similarity measures for mould matching. Johnson et al.~\cite{Johnson2015} use radiography images of Rembrandt's prints for chain line detection. Their method uses the Radon transform to estimate the orientation of chain lines and a combination of vertical filtering and Hough transform to extract the lines. This forms the basis of their mould matching algorithm. Shifting gears to transmitted light images for chain line extraction, Hiary et al.~\cite{Hiary2007} developed a model to separate background and foreground and then used the Radon transform to detect chain lines. They use canny edge detection to additionally extract the watermark. Most recently, both Biendl et al.~\cite{Biendl2021} and Sindel et al.~\cite{Sindel2021} developed deep learning approaches for chain line segmentation and parametrisation from transmitted light images. While~\cite{Biendl2021} follow a supervised approach to train a UNet for chain line segmentation,~\cite{Sindel2021} trained a conditional generative adversarial network to predict a segmentation mask. However, both approaches rely on supervised training which in turn requires manually annotated ground truth segmentation masks and is therefore costly. The work by Xi et al.~\cite{Xi2016} additionally showed that mould matching using chain line information alone is not enough to successfully identify different paper stock. Considerably less work has been published on the extraction of laid lines. Atanasiu et al.~\cite{Atanasiu2002} first uses emboss filtering for noise removal and subsequently applied the discrete fast Fourier transform to obtain the highest frequency in the Fourier domain representing the frequency of laid lines. More recently, Gorske et al.~\cite{Gorske2021} proposed a laid line density approximation based on the 1D fast Fourier transform and refinement of the frequency estimate through a phase vocoder. While they do not produce a paper average number of laid lines per centimetre, as commonly used in laid line analysis, their results show a map of local laid line densities for the full image and enable paper matching based on the similarity of density distributions throughout the image. 

Considering the vast amount of reflected light images readily available from many archives and libraries, there is a need to broaden the study of the evidence provided by the mould. The use of automatic feature extraction from reflected light images has the potential to improve the speed, scale and accuracy of paper stock identification, facilitate research into heritage questions we can only answer at scale and transform this field of study. Without the need for expensive equipment or re-imaging of paper manuscripts, paper stock identification becomes accessible even for small institutions. 

In this paper, we propose analysing and extracting paper features from reflected light images by means of mathematical multi-scale image decomposition methods (cf. Figure \ref{fig.abstract}). We show that the spectral total variation (TV) decomposition~\cite{Gilboa2013,Gilboa2014} is able to separate the features of interest (chain and laid lines) from ink and noise. In a sense, this can be seen as the simulation of transmitted light images from reflected light. Furthermore, we develop a semi-automated approach for the extraction and measurement of chain and laid lines from the decomposed images. We identify straight lines and count their number and density to identify paper stock signatures. In that, we demonstrate the feasibility of reflected light images for paper feature analysis. To the best of our knowledge, the application of the proposed method to reflected light images has not been attempted before. In what follows, we show how we have developed mathematical imaging approaches for extracting granular paper feature information from regular reflected light images of paper manuscripts. The accompanying code is available at \url{https://github.com/TamaraGrossmann/HiddenKnowledge}.

\subsection{Organisation} The paper is structured as follows: First, we introduce the spectral total variation decomposition that forms the basis of our method in the second section. We present the proposed algorithms for chain and laid line extraction. In the third section "Results and Discussion", we show results of the line extraction on example images. Additionally, we discuss the selection of parameters in the semi-automated algorithms. In the last section we draw conclusions and outline future work.

\section{Methods}
The analysis of medieval paper features from reflected light images is challenging for three main reasons. First, the features of interest, the chain and laid lines, have low contrast in the image and therefore can be barely visible and hard to detect. Additionally, many papers contain writing with ink occluding or interrupting paper features. Finally, medieval manuscripts tend to show varying levels of paper degradation, dirt spots and image noise that interfere with the line extraction. For these reasons, the direct chain and laid line extraction on reflected light images is not feasible. Instead, we apply spectral total variation decomposition to separate the granular, low contrast features of interest from the rest of the reflected light image.

\subsection{Pre-processing}
In order to produce absolute measurements from the chain and laid line extraction in terms of distances in millimetre and line density per centimetre, the first step is the measurement of the pixel size. For reflected light images, we typically observe two types of data. That is, either the image contains a ruler in one of the corners or the paper size (height and width) is provided in the meta data. Using the paper measurements to extract the pixel size can introduce some inaccuracies due to paper being, for example, warped on the imaging platform. However, these inaccuracies are minor and we have found them to not obscure the absolute measurements of line distances and densities. Both types of data require the detection of edges of either the ruler lines or the edges of the paper itself. We employ canny edge detection~\cite{Canny1986}, a well-established method that is based on the image intensity gradient and involves Gaussian smoothing and thresholding. In the case of a ruler being present in the image, the size of each pixel can be determined by taking the pixel distance between two ruler lines. When using the meta data information on the paper size, we determine the pixel size by measuring the pixel distance between the top and bottom edge of the paper. The final step in pre-processing is transforming the images into greyscale for simplicity. 

\begin{figure*}[ht]
    \centering
    \includegraphics[width=0.95\textwidth]{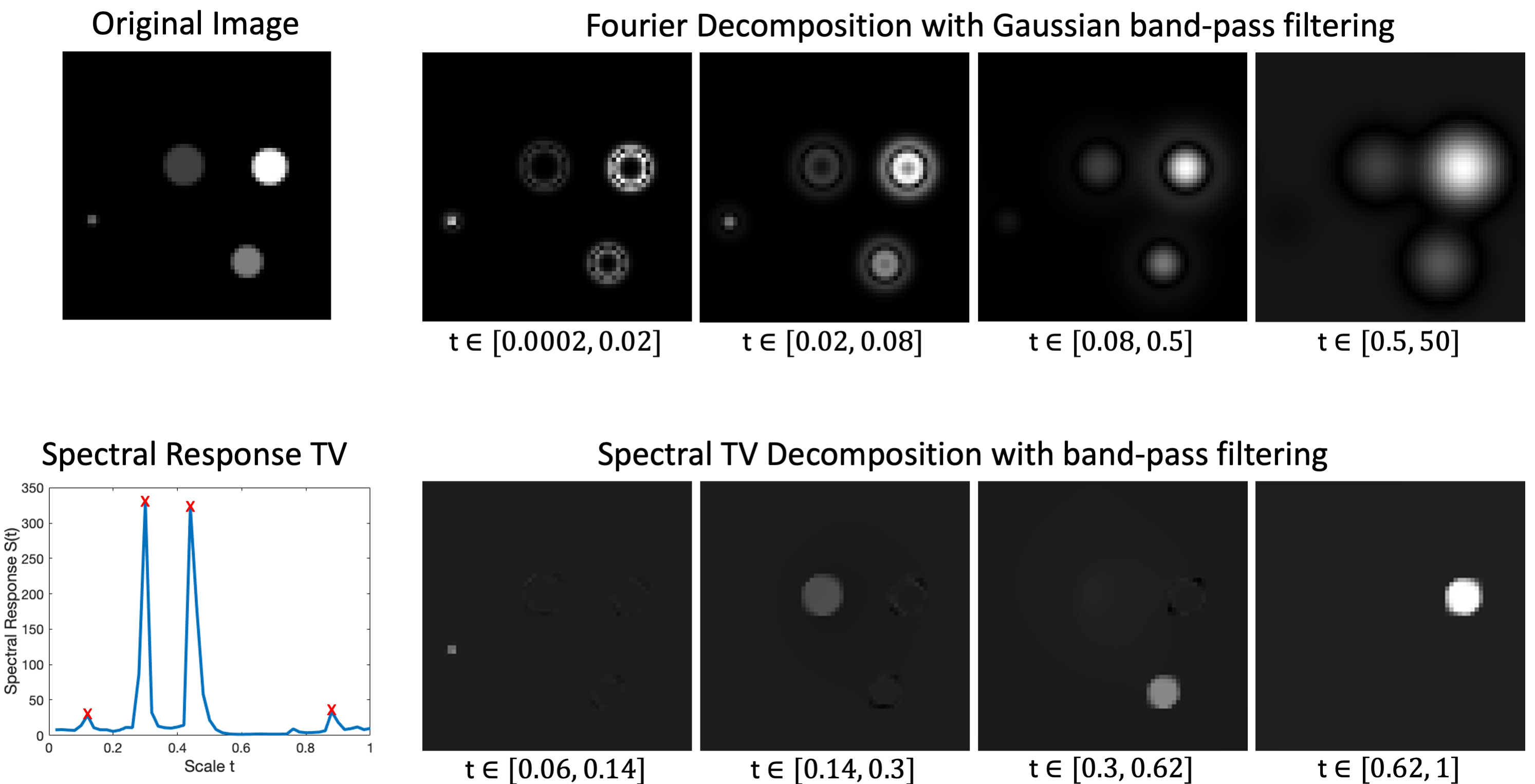}
    \caption{Example of Fourier decomposition with Gaussian band-pass filtering (top row) in comparison to spectral TV decomposition with band-pass filtering (bottom row) for an image containing four disks at different sizes and contrasts (original image on the left). Images represent high to low frequencies/ small to large scales from left to right. For the spectral TV decomposition, the spectral response is plotted (bottom row), the red crosses mark the spectral responses of the individual disks. Note that to aid visualisation in the Fourier decomposition each image has been rescaled independently.}
    \label{fig.disk_decomp}
\end{figure*}
\subsection{Spectral Total Variation Decomposition}
The decomposition of an image into its scale-dependent components is a common and successful tool in image analysis. The intuitive idea is to separate components loosely related to their size/scale. Exact representations can be modelled by mathematical approaches relating to different types of components. A prominent example is the Fourier transform that decomposes an image based on the frequency of its structures. The basis of this decomposition are trigonometric functions. Especially for data with repeating structures, such as signals made up of multiple waves with varying frequencies, the Fourier transform is a powerful tool to separate the different frequency components. By defining filters, the structures can be separated in the transform domain relating to scale/frequency. One such filter is, for example, the Gaussian filter. The standard deviation $\sigma$ of the Gaussian filter determines the frequencies that are attenuated or removed in the Fourier domain. The resulting images depict varying degrees of blurring. In a two-dimensional setting, Gaussian filtering in the Fourier domain of an image $f(x)$ is defined as $\mathcal{F}[G_{\sigma}](w)\mathcal{F}[f](w)$, where $\mathcal{F}$ is the Fourier transform and $G_{\sigma}(x) = \frac{1}{2 \pi \sigma^2}\exp{(-\lvert x \rvert^2/2\sigma^2)}$ the Gaussian filter kernel. This Gaussian filtering can be equivalently modelled by the solution to the heat equation $u_t(x,t) = \Delta u(x,t)$ with the initial condition $u(x,0) = f(x)$ and for $t = 2 \sigma^2$. It is also called the Gaussian scale-space representation of $f(x)$. The temporal component of the heat equation is related to the standard deviation of the Gaussian kernel and describes the scale in the decomposition. That is, the solution of the heat equation at larger time instances equates to Gaussian filtering in Fourier domain with higher standard deviations and therefore higher degrees of blurring. An example of Fourier decomposition with Gaussian band-pass filtering at different frequencies is shown in Figure \ref{fig.disk_decomp}. Linear scale-space representations of an image, such as for Gaussian filtering, however, are not always the best choice for decomposition as images are inherently non-linear. Fourier decomposition can approximate edges of an image only at high frequencies, as seen in the Gaussian filtered example in Figure \ref{fig.disk_decomp}. Edges define the boundary of objects in an image and a representation of these features only at high frequencies can therefore make the clean separation of image structures challenging. Instead, the spectral total variation decomposition as introduced by \cite{Gilboa2013,Gilboa2014} is based on the non-linear edge-preserving total variation functional and retains sharp edges at all scales. The spectral TV decomposition is driven by the size and the contrast of the structures with small and low contrast features being represented at lower scales and large, high contrast features at higher scales. The basic atoms are induced by eigenfunctions to the subdifferential of the TV functional. These are, for example, disks and disk-like structures. The spectral TV decomposition for an image containing disks is shown in Figure \ref{fig.disk_decomp}. In contrast to the Fourier decomposition, the spectral TV decomposition is able to clearly separate each of the disks while retaining sharp edges. The relation between structure size and contrast is particularly relevant for the analysis of paper images. In reflected light images, the chain and laid lines are small features and only present at low contrast while the noise, dirt spots, paper degradation such as water damage, and ink are larger features or are present at higher contrast. The spectral TV decomposition is therefore ideal to separate the structures of interest (chain and laid lines) from any components that might impede the feature analysis of paper. 

\begin{figure*}[ht]
    \centering
    \includegraphics[width=0.95\textwidth]{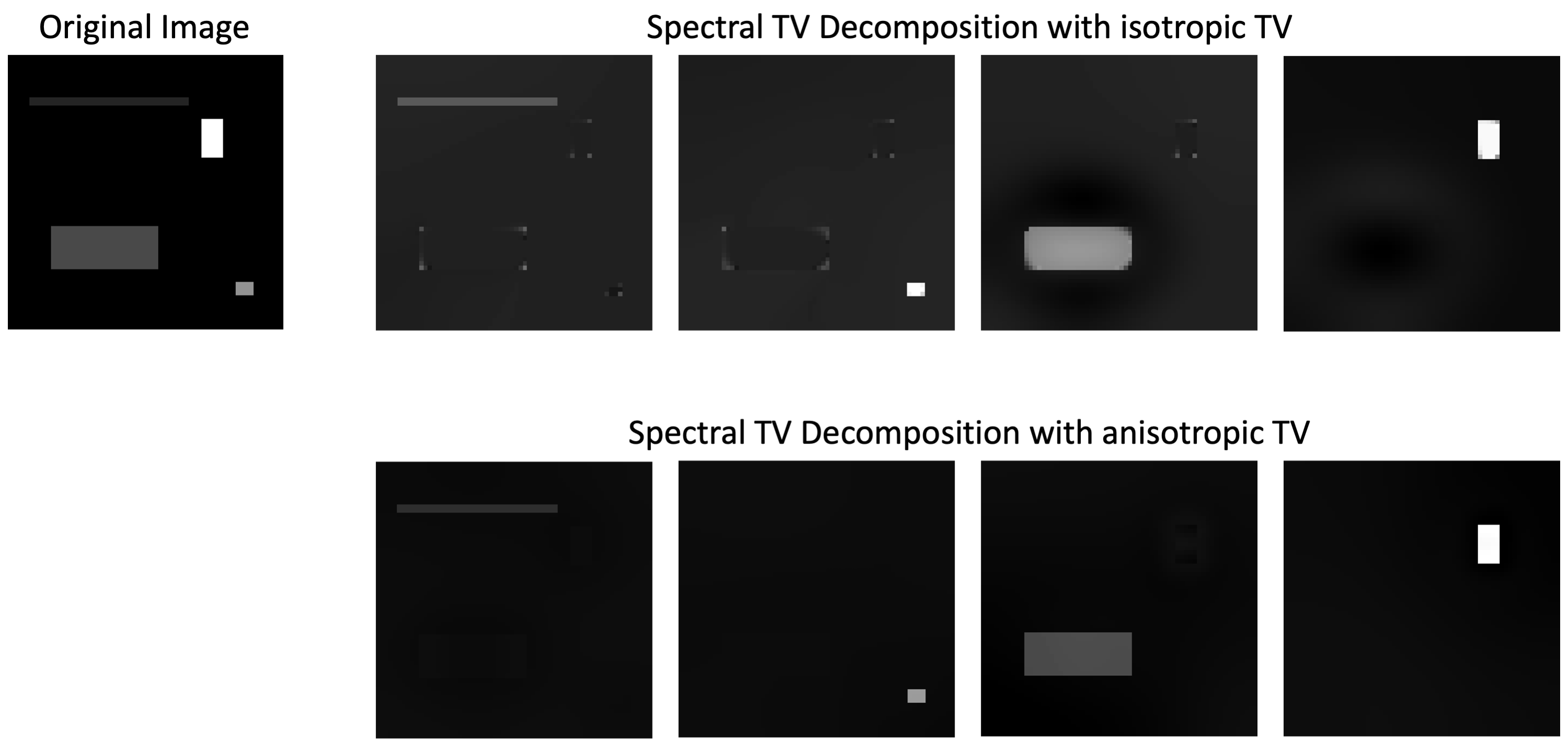}
    \caption{Example of spectral TV decomposition with isotropic (top row) and anisotropic (bottom row) TV on an image containing rectangles of different size and contrast.}
    \label{fig.rect_decomp}
\end{figure*}

The total variation functional is defined as $J_{TV}(u) = \int_{\Omega} \lvert D u \rvert$ with $\Omega$ being the image domain and $Du$ the distributional gradient. For $u$ smooth enough, this becomes $J_{TV}(u) = \int_{\Omega} \lvert \nabla u \rvert dx$~\cite{Aubert2002}. The TV scale-space representation of an image $f(x)$ is modelled by the TV flow with Neumann boundary conditions, formally written as~\cite{Andreu2001,Bredies2016}:
\begin{align}\label{eq.TVflow}
    \begin{cases}
    u_t(t,x) &= \text{div}\left( \frac{D u}{\lvert D u \rvert}\right) \quad t \in (0, \infty), \,\, x \in \Omega\\
    u(0,x) &= f(x) \quad x \in \Omega\\
    \frac{\partial u}{\partial \nu}(t,x) &= 0, \quad t \in (0, \infty), \,\, x \in \partial \Omega,
    \end{cases}
\end{align}
where $\Omega$ is the image domain and $\partial \Omega$ denotes the boundary of the image domain. The spectral TV decomposition is based on the solution $u(t,x)$ to the TV flow \eqref{eq.TVflow} given an image $f(x)$. The TV transform is then derived from the second temporal derivative of the TV flow solution \cite{Gilboa2013,Gilboa2014}:
\begin{align*}
    \phi(t,x) = u_{tt}(t,x)t,
\end{align*}
where $t$ refers to the scale parameter. The transform generates impulses at the basic atoms or structures. To give an intuition, in the case of an image containing disks the TV transform will create an impulse at a scale based on the radius and contrast of the disk. Disks of different radii and contrasts will therefore result in impulses at different scales $t$ and enable the separation of the disks. One such example is shown in Figure \ref{fig.disk_decomp}. Notice the four peaks or impulses in the spectral response plot that are related to the scales at which the disks are separated. Given the spectral responses $\phi (t,x)$, the initial image can be recovered via the inverse TV transform $f(x) = \int_{0}^{\infty} \phi(t,x) dt + \bar{f}$, with $\bar{f}$ the mean value of $f$. Instead of recovering the full images, we can define filters and manipulate the image in the transform domain, similar to Gaussian filtering in the Fourier domain. In the following, we will be using band-pass filtering, i.e., separating the image based on specific scales. The TV-band-pass filtering in an interval $[t_k, t_{k+1}]$ for $t_k < t_{k+1}$ is defined as
\begin{align*}
    b^k &= \int_{t_k}^{t_{k+1}} \phi(t,x) dt \quad \text{for} \quad k = 1,\dots, K-1 \\ 
    b^K &= \int_{t_{K-1}}^{t_{K}} \phi(t,x) dt + f_r(t_K,x).
\end{align*}
The filtered images $b_k$ are referred to as the $K$ decomposed spectral bands of the initial image $f$ and the residual is derived as $f_r(t_K,x) = u(t_K,x)-u_t(t_K,x)t_K$. The spectral bands sum up to the initial image $f = \sum_{k = 1}^{K} b_k$. For the numerical derivation of the TV flow solution and the TV decomposition we refer to the original papers by Gilboa~\cite{Gilboa2013,Gilboa2014}. Additionally,~\cite{Grossmann2020, Grossmann2022} have developed deep learning approaches for fast computations. Code is openly available for both methods \footnote{Code for the model-based approach by Gilboa et al. is available at \url{https://guygilboa.net.technion.ac.il/2020/10/09/spectral-total-variation-color/}. For the deep learning approach by Grossmann et al., code is available at \url{https://github.com/TamaraGrossmann/TVspecNET}.}. 

\begin{figure*}[t]
    \centering
    \includegraphics[width=0.95\textwidth]{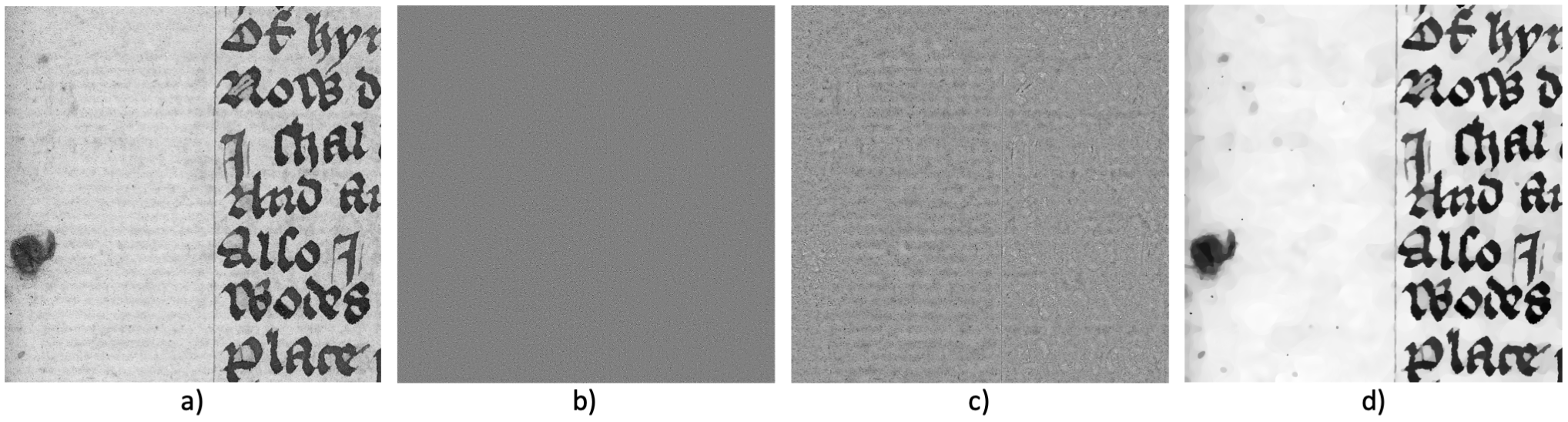}
    \caption{Example of spectral TV decomposition of a reflected light image - patch taken from Cambridge University Library MS Ii.5.41, fol. 10r. a) the original image, b) high-pass filtered image containing only image noise, c) band-pass filtered image including the lines of interest, d) low-pass filtered image displaying larger and high contrast structures such as the ink and dirt spot. Reproduced by kind permission of the Syndics of Cambridge University Library, MS Ii.5.41, fol. 10r. Full paper image is displayed in Figure \ref{fig.laidline_extraction}a).}
    \label{fig.decomposition_example}
\end{figure*}

The classical spectral TV decomposition as introduced above is based on the isotropic total variation, i.e. the discrete TV functional is rotation invariant and in the two-dimensional setting defined as 
\begin{align*}
    J_{TV}^{iso}(u) = \sum_{i,j}\sqrt{u_{x_1}^2(i,j) + u_{x_2}^2(i,j)},
\end{align*}
where $\nabla u = (u_{x_1},u_{x_2})$ denotes the derivatives in $x=(x_1,x_2)$ direction respectively and $i,j$ the pixel locations. The eigenfunctions or basic atoms related to isotropic TV are therefore disk-like shapes as described above. However, we can also discretise the TV functional to include directionality in x- and y-direction by using the 1-norm to derive the gradient magnitude, i.e. 
\begin{align*}
    J_{TV}^{ani}(u) = \sum_{i,j}\lvert u_{x_1}(i,j) \rvert + \lvert u_{x_2}(i,j)\rvert.
\end{align*}
Let us call this discretisation anisotropic TV. It is not rotation invariant and the related eigenfunctions are rectangular shapes. An example of the effect of isotropic versus anisotropic TV decomposition of an image containing rectangles is shown in Figure \ref{fig.rect_decomp}. The corners of the rectangles are recovered as sharp edges in the anisotropic case, however, the isotropic TV decomposition shows rounded corners of the rectangles and noise at the corner locations in the other bands. While we will mostly use the classical spectral TV decomposition with isotropic TV for paper feature analysis, we will show in the results section that there are cases in which anisotropic TV will be more beneficial.

We apply the spectral TV decomposition to the reflected light images of paper and use band-pass filtering to separate the chain and laid lines from the rest of the paper. In a sense, this decomposition can be seen as a simulation of transmitted light images as it creates images that show similar clarity of the lines to transmitted light. In Figure \ref{fig.decomposition_example}, an example of the spectral TV decomposition for paper is shown. To separate the relevant features, the spectral bands need to be chosen appropriately. Image noise is typically very small and will be present in the lowest scales, i.e., around $t=0$. The chain and laid lines are slightly larger structures with low contrast and are represented in low scales. In contrast, the ink has a high contrast and can therefore be found at the higher bands of the decomposition. We have found the best results for extracting the chain and laid lines by selecting the band-pass filter interval around $t \in [0.026, 1]$. We will discuss the selection of spectral bands in more detail in the results section. 

\subsection{Chain Line Extraction}
Chain line imprints on medieval paper are produced by the vertical chain wires in the mould. As they sit on top of the laid wires, the imprint tends to be more visible than laid lines in reflected light images and casts a shadow in transmitted light images. On average, chain lines are placed at a distance of between 1.5 - 5cm~\cite{vanstaalduinen2006}. Around a watermark, the chain line distances can be reduced as the vertical wires act as supports onto which the watermark is fixed. A variation in the chain line distance across a paper can therefore give an indication as to where the watermark is placed. To this end, we are interested in quantifying the distance between each chain line across the entire page for extracting the fingerprint of the paper. While the average distance between chain lines can give relevant insights, more detailed information and precise matching can be performed when taking distance variations across a paper into account and using the entire chain line sequence. This process gives an overall idea of the size of the mould and its construction.

\begin{algorithm}[t]
\caption{Chain Line Extraction}\label{alg.chainline}
\begin{algorithmic}[1]
\vspace{2pt}
\State \textbf{Pre-processing:} Measurement of pixel size in relation to paper size.
\State \textbf{Patch selection:} Crop image to smaller area contain all chain lines.
\State \textbf{Spectral TV decomposition:} Separate small, low contrast features from high contrast structures such as ink.
\State \textbf{Fourier filtering:} Enhance chain lines in frequency domain.
\State \textbf{Projection:} Project filtered image onto relevant axis.
\State \textbf{Peak detection:} Locate peaks with highest magnitude, representing the chain lines.
\State \textbf{Chain Line distance measurement and plotting}.
\vspace{2pt}
\end{algorithmic}
\end{algorithm}

\begin{figure*}[t]
  \centering
  \includegraphics[width = 0.99\textwidth]{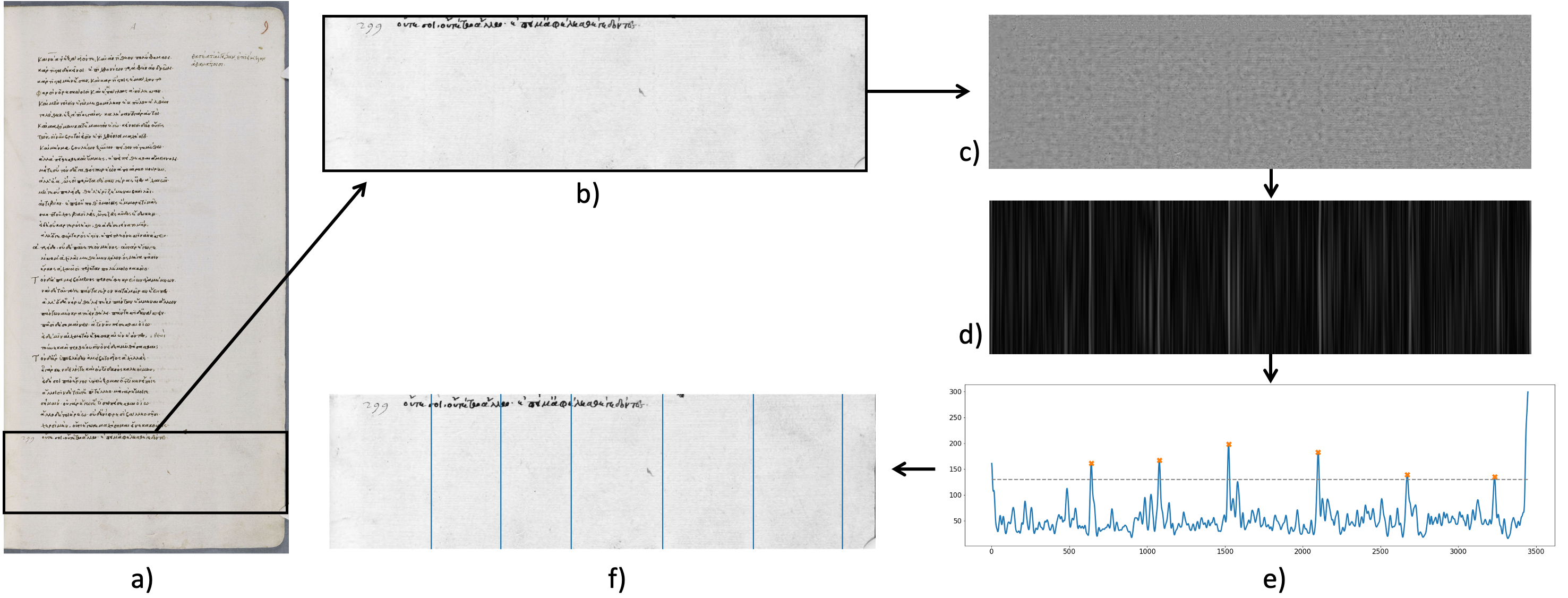}
  \caption{Chain line extraction pipeline: a) original colour image with patch marked; b) grayscale patch; c) spectral TV band-pass filtered image; d) Fourier filtered image; e) plot of d) projected onto x-axis, orange crosses mark the peaks and the grey dotted line represents the threshold; f) original image with detected lines. Original image courtesy of The Parker Library, Cambridge, Corpus Christi College, MS 081: Homer, Iliad, Odyssey. Quintus of Smyrna, Posthomerica, p. 9.}
  \label{fig.chainline_extraction}
\end{figure*}

The first step in our chain line extraction pipeline is the selection of a patch of the image to reduce computational cost. Spectral TV decomposition can become computationally expensive especially when considering high resolution images such as those available for paper. However, in order to get the chain line distances we do not need to analyse the entire page, but rather a subsection. Two main features need to be considered in the selection of patches: Depending on the size and cut of the paper, e.g., folio or bi-folio, the orientation of the chain lines can appear rotated compared to the mould such that they are horizontal instead of vertical. Additionally, as we aim to extract the sequence of chain line distances, the patch needs to contain every line at least partially. That is, we do not need the full length of the line to extract the distances. An example for patch selection is shown in Figure \ref{fig.chainline_extraction}a). In this case, the chain lines are vertical and we select a patch in the lower quarter of the paper page. We assume the chain line orientation to be either horizontal or vertical. For images with lines at different orientation, the Radon transform can be used to determine the dominating orientation and rotate the image accordingly to have chain lines presented horizontally or vertically, compare e.g.,~\cite{Johnson2015}. After patch selection, the next step is computing the spectral TV decomposition of the patch as introduced above. The band-pass filtered image forms the base for the detection of chain lines. We are left with an image that now only contains the chain and laid lines and any structures that interfere with the line detection and extraction are removed (cf. Figure \ref{fig.chainline_extraction}c)). We assume the chain lines to be straight and parallel to each other and employ Fourier filtering to detect the lines. While we have argued above that Fourier decomposition is not ideal to separate small, low contrast structures from high contrast features in images in general, it is suited for periodic structures. Therefore, the task at hand lends itself to Fourier filtering. In other words, as the chain lines in the images are present with relative regularity, using a frequency-based method can enhance and detect the lines. We use the fast Fourier transform on the band-pass filtered image. In the Fourier domain, repetitive structures are presented as peaks along the axis orthogonal to their orientation in the image with the distance to the centre marking the frequency. For chain lines that lie vertically in the image, the lines will generate a peak along the horizontal line in Fourier domain. We employ rectangular filtering similar to \cite{Johnson2015} along the relevant axis to enhance the lines and remove any other structures. The filter width is 1-3 pixels and the height is determined by the image height. The resulting filtered image (cf. Figure \ref{fig.chainline_extraction}d)) displays only the lines in the relevant direction with the chain lines enhanced. Due to noise, however, additional lines will be present with lower intensity. The next step is therefore to project the image onto the axis perpendicular to the chain lines and employ peak detection. The peaks with the highest magnitude correspond to the chain lines, while the noisy lines will only generate low magnitude peaks. Gaussian filtering of the 1D signal removes some of the noise. We perform a simple peak detection by comparing neighbouring values and thresholding to filter out any remaining noise (cf. Figure \ref{fig.chainline_extraction}e)). The threshold is manually selected. Once the chain line peaks are located, we can determine their pixel distance. Using the pixel measurements from the pre-processing step, the pixel distances can then be transformed into metric values. Finally, we draw the detected chain lines in the original image for visualisation as shown in Figure \ref{fig.chainline_extraction}f). 

\begin{figure*}[t] 
  \centering
  \includegraphics[width = 0.98\textwidth]{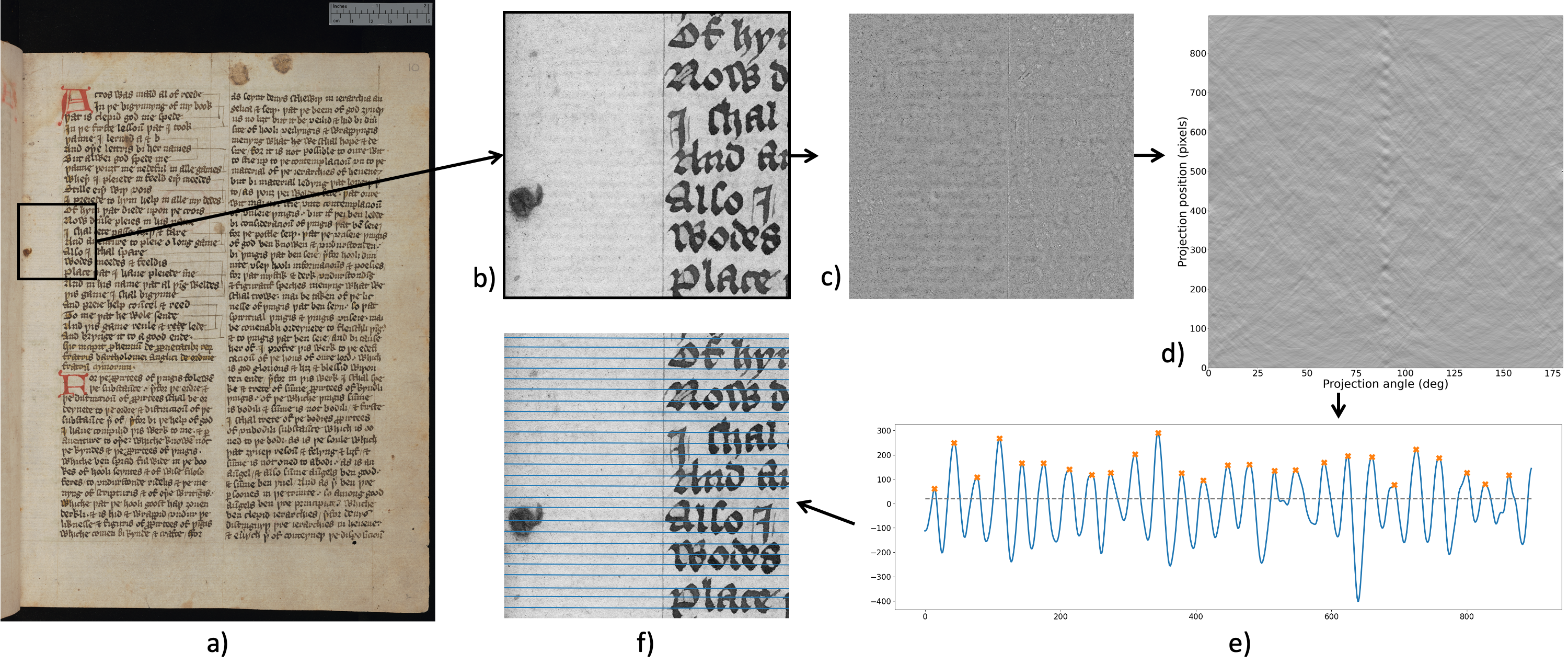}
  \caption{Laid line extraction pipeline: a) original colour image with patch marked; b) grayscale patch; c) spectral TV band-pass filtered image; d) Radon transform image depicting projection angles between 0° and 180° and the projection position; e) plot line through image d) at the 90° angle, orange crosses mark the peaks and the grey dotted line represents the threshold; f) original image with detected lines. Reproduced by kind permission of the Syndics of Cambridge University Library, MS Ii.5.41, fol. 10r.}
  \label{fig.laidline_extraction}
\end{figure*}
\subsection{Laid Line Extraction} \label{sec.laid}
The laid lines on medieval paper are imprints from the horizontal wires in the mould. They are densely spaced and appear at low intensity in reflected light images. The laid line density is measured as lines per 1cm. On average, medieval paper will have laid line densities of 5-15 lines per centimetre~\cite{vanstaalduinen2006}. While the distances between laid lines can also be of interest, we will focus on the laid line density in this paper. Note that laid line distances can be computed from the extraction results. Our approach is, however, not tailored to have the extracted lines placed in the centre of the laid lines and distance measurements between lines can therefore be slightly imprecise.

\begin{algorithm}[t]
\caption{Laid Line Extraction}\label{alg.laidline}
\begin{algorithmic}[1]
\vspace{2pt}
\State \textbf{Pre-processing:} Measurement of pixel size in relation to paper size.
\State \textbf{Patch selection:} Crop image to smaller square area.
\State \textbf{Spectral TV decomposition:} Separate small, low contrast features from high contrast structures such as ink.
\State \textbf{Radon transform:} Detect laid lines in the Radon domain.
\State \textbf{Cross-section:} Take a cross-section along the relevant angle to obtain line peaks.
\State \textbf{Peak detection:} Locate peaks with highest magnitude, representing the laid lines.
\State \textbf{Laid Line density measurement and plotting}.
\vspace{2pt}
\end{algorithmic}
\end{algorithm}

Similar to chain line extraction, the first step is the selection of a patch on which the laid line extraction is implemented. As the aim is to obtain the density of laid lines per centimetre, the patch needs to be large enough to contain 1cm of the paper sheet. We select square patches at places of the paper that have the best visibility of the laid lines. These can often be found near the inner hinge of the book. Next, the spectral TV decomposition of the patch is computed. The resulting band-pass filtered image only contains the chain and laid lines (cf. Figure \ref{fig.laidline_extraction}c)). While for the chain line extraction Fourier filtering with a rectangular filter mask was sufficient to extract the lines, in reflected light images the method fails for laid lines. Laid lines tend to be less visible than chain lines due to their placement in the mould. Therefore, the band-pass filtered images remain too noisy to define an appropriate filter mask to extract the densely spaced lines in the Fourier domain. Instead, we propose the use of the Radon transform, similar to~\cite{Hiary2007,Xi2016} only for laid line detection. The Radon transform computes parallel line integrals through the image at different angles. That is, pixel values of the image are projected along lines rotated around the image centre. Lines in the images are then represented as peaks in the Radon domain (cf. Figure \ref{fig.laidline_extraction}d)). The resulting image depicts the line integral value for each projection angle and projection position related to the image centre. The location of each laid line and the angle at which it is present can then be extracted. To this end, we determine the projection angle with the highest magnitude to obtain the orientation of the laid lines and plot the projection positions along that angle (cf. Figure \ref{fig.laidline_extraction}e)). Based on the resulting 1D signal, we perform peak detection. Note that we do not use any filtering in the transform domain or inverse transform as we did in the chain line extraction. However, there is a distinct relationship between the Radon and the Fourier transform through the Fourier slice theorem~\cite{Bracewell2004}. More specifically, the 2D Fourier transform of an image along a line at angle $\alpha$ is equivalent to taking the 1D Fourier transform of the Radon transform at angle $\alpha$ for the same image. Essentially, we leverage the Fourier transform for both the chain and laid line extraction. While the chain line extraction looks at angles of only 0° and  90° and uses rectangular filtering, the laid line extraction takes all angles into account and instead of filtering, the line location is directly extracted from the transform domain. Therefore, the basic idea for the line detection is similar. The 1D signal in the laid line extraction is smoothed using a Gaussian kernel to remove any remaining noise. We then perform the same peak detection approach and thresholding as in the chain line detection. As the peaks in the 1D signal contain the location of the laid lines, we can count the lines per centimetre and draw the detected laid lines in the original image for visualisation (cf. Figure \ref{fig.laidline_extraction}f)).

Code for both chain and laid line extraction is available at \url{https://github.com/TamaraGrossmann/HiddenKnowledge}.

\section{Results and Discussion} \label{sec.results}
In this section, we demonstrate the performance of the proposed algorithms for chain and laid line detection on a set of reflected light images of paper manuscripts from The Parker Library, Corpus Christi College, Cambridge, UK. We additionally discuss the parameter selection in the semi-automatic line extraction algorithms.

\begin{figure*}[t] 
  \centering
  \includegraphics[width = 0.98\textwidth]{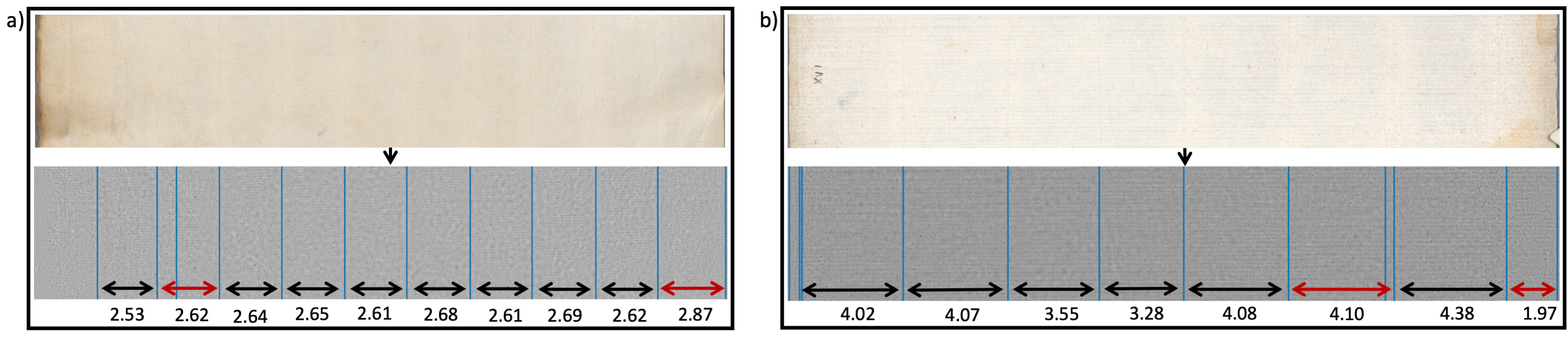}
  \caption{Results for chain line extraction of two different papers, the top images show the reflected light image and the bottom images the spectral TV decomposed images with the detected lines drawn. a) Patch taken from The Parker Library, Cambridge, Corpus Christi College, MS 171A: Scotichronicon (Volume 1), p. iiv. The average chain line distance is 2.63cm. Individual chain line distances in cm are shown in the image. The algorithm detected two additional lines, marked by red arrows. b) Patch taken from The Parker Library, Cambridge, Corpus Christi College, MS 171A: Scotichronicon (Volume 1), p. xvir. The average chain line distance is 3.93cm. Individual chain line distances in cm are shown in the image. The algorithm detected two additional lines, marked by red arrows.}
  \label{fig.chainline_result}
\end{figure*} 
\subsection{Chain Line Extraction}
For chain line extraction, images were processed as described in Algorithm \ref{alg.chainline}. For two example images, the results are shown in Figure \ref{fig.chainline_result}. We are able to successfully extract the chain lines and measure the distances between all detected lines. Both of the example images show discolouration along the edges of the paper. These types of paper degradation are not uncommon in medieval paper. The spectral TV decomposition is, however, able to separate the discoloured, darker patches from the lighter and finer lines in the image enabling chain line detection without interference of noise. In Figure \ref{fig.chainline_result}a), a line is visible on the far left of the paper sheet that has not been detected by the algorithm. This line might or might not be a chain line, or more likely a fold of the page. Our proposed method therefore is able to distinguish between different types of lines even with slight contrast differences. In Figure \ref{fig.chainline_result}b), two of the chain line distances are significantly below the mean and therefore give an indication on the position of the watermark. Further inspection of the full page confirms the finding.
On the other hand, for both of the example images shown in Figure \ref{fig.chainline_result}, the algorithm detects additional lines that are not in fact chain lines but rather the result of noise in the paper. This can be rectified in two ways. Either the algorithm is re-run with changed parameters (see below for parameter selection), or by making an informed decision on which lines are to be omitted in the distance measurement. The edge of the paper is one of the lines that the algorithm may detect additionally. However, those lines can easily be omitted by the user.

The two examples highlight the feasibility of the proposed algorithm for chain line detection even for paper with degraded quality and noise.

\begin{figure*}[t] 
  \centering
  \includegraphics[width = 0.98\textwidth]{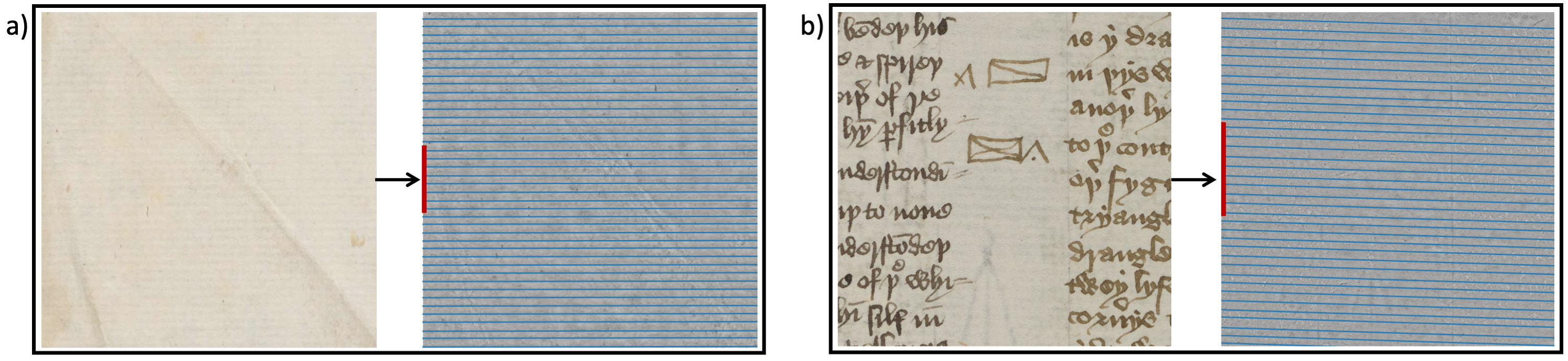}
  \caption{Results for laid line extraction of two different papers, the left images in each box show the reflected light image and the right images the spectral TV decomposed images with the detected lines drawn. a) Patch taken from The Parker Library, Cambridge, Corpus Christi College, MS 100: Transcripts (16th century). Simeon of Durham OSB. Geoffrey of Coldingham OSB. Tito Livio Frulovisi. Walter of Guisborough OSA. Asser, p. iiv. The laid line density measured in the 1cm section marked by the red line is 8 lines per cm.  b) Detail from Cambridge University Library, MS Ii.5.41, fol. 336r. Reproduced by kind permission of the Syndics of Cambridge University Library. The laid line density measured in the 1cm section marked by the red line is 10 lines per cm.}
  \label{fig.laidline_result}
\end{figure*}
\subsection{Laid Line Extraction}
For laid line extraction, we process images as described in Algorithm \ref{alg.laidline}. Results for two example images are shown in Figure \ref{fig.laidline_result}. The proposed algorithm is able to successfully extract all laid lines of the paper patch and obtain the laid line density as the number of lines in centimetre. The paper shown in Figure \ref{fig.laidline_result}a) has a laid line density of 8 lines per centimetre. Despite the two large folds across the paper patch that interrupt the laid lines, our algorithm is able to detect all laid lines. The folds have a higher contrast and the spectral TV decomposition can therefore separate those structures from the laid lines. Additionally, the folds are present at a different angle to the lines of interest and the plot of the Radon transform at the laid line angle will not pick up the fold. In Figure \ref{fig.laidline_result}b), the laid lines are obstructed by a large amount of ink. However, due to the clear difference in contrast and size, the spectral TV decomposition is able to separate ink and paper. This example demonstrates that our method detects the laid lines even in this highly obstructed case. The resulting density is 10 lines per centimetre.

\begin{figure*}[t] 
  \centering
  \includegraphics[width = 0.98\textwidth]{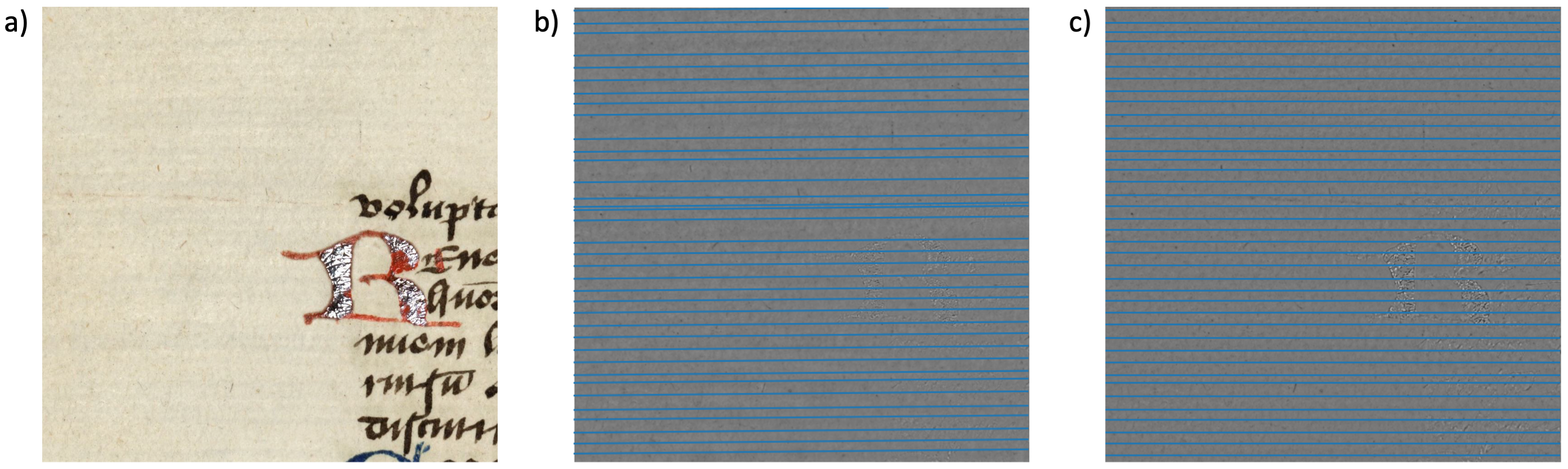}
  \caption{Comparison of laid line extraction with a) isotropic TV decomposition and b) anisotropic TV decomposition. Patch taken from The Parker Library, Cambridge, Corpus Christi College, MS 078: Domenico di Bandino, Fons memorabilium uniuersi (book 5, De uiris claris), p. 3.}
  \label{fig.laidline_anisotropic}
\end{figure*}

The extraction of laid lines tends to be more challenging than the extraction of chain lines. The imprint of the chain wires is often more visible. This is due to the chain wires sitting on top of the laid wires. Additionally, the laid lines are spaced densely resulting in noise having a bigger influence on the results. For paper or images with reduced quality, this can lead to lines not being detected with the proposed algorithm that uses isotropic TV in the decomposition. An example is shown in Figure \ref{fig.laidline_anisotropic}b). Instead, we propose anisotropic TV for spectral TV decomposition in cases where isotropic TV decomposition fails. As introduced in the method section above and visualised in Figure \ref{fig.rect_decomp}, the basic atoms of anisotropic TV decomposition are rectangular shapes. The evolution of the TV flow \eqref{eq.TVflow} for anisotropic TV will result in more block-shaped structures. We leverage this in cases of high noise levels or low image quality to be able to extract the laid lines. As shown in Figure \ref{fig.laidline_anisotropic}, we are then able to extract all laid lines despite lower image quality.

\subsection{Parameter Selection}
The proposed algorithms for chain and laid line extraction are semi-automated and some parameters need to be manually tuned and adjusted heuristically for each paper image depending on the image resolution, quality and lines present. These parameters are the interval for band-pass filtering of the spectral TV decomposition and the smoothing factor and thresholding for peak detection. For chain line extraction, the width of the filtering mask for Fourier filtering needs to be additionally selected. 

The band-pass filtering interval is related to the image resolution. As the decomposition is dependent on the size of the structures and lines in images with higher resolution will typically be represented by a larger number of pixels, the scale at which they appear in the scale-space representation will be larger. Conversely, images with a lower resolution will need band-pass filtering at lower scales. For the Corpus Christi manuscripts, we have found that $t \in [0.026,1]$ for band-pass filtering gives the best results. This selection ensures that small image noise is filtered out but all line features are retained. However, the algorithm is robust to small changes in the interval and will detect correct lines with varied bands as long as the line information is contained in the spectral bands. In chain line extraction, the second parameter is the filtering mask for Fourier filtering. Johnson et al.~\cite{Johnson2015} suggested a rectangular mask of 1/3 image height and 3px width. This parameter is also dependent on the image resolution and the number of pixels that the lines are represented by. For the example images shown, a rectangular mask with 2/3 image height and 1px width has given the best results, where the choice of width has greater importance. Finally, for both extraction algorithms the smoothing factor and threshold for peak detection is to be determined. This parameter is dependent on the image noise and degradation. Images with higher noise level require a larger smoothing factor. In the algorithms presented, the threshold is chosen manually, however, there are approaches to automate the peak detection and thresholding~\cite{Rabbani2011}. This will be part of future work.

\section{Conclusions}
The extraction of the fingerprint of medieval paper has been at the core of much research and for many scholars. In this paper, we proposed algorithms to detect and extract the chain and laid lines from reflected light images. In that, we showed the feasibility of leveraging the so far untapped resource of reflected light images for paper analysis where most other work has been focused on radiography or transmitted light images. 

We introduced the use of spectral TV decomposition for separating high contrast and large size structures such as ink and discolouration from the low contrast and fine structures that are the lines of interest. Subsequently, we proposed algorithms for line extraction based on the decomposed images. The results clearly demonstrate the successful detection and extraction of all chain and laid lines even in cases of paper degradation such as discolouration or folds. We additionally showed an alternative decomposition approach for cases in which paper or image quality made the detection more challenging.

This proof of concept demonstrates that mathematical analysis can be used to extract features of the imprint that the mould left on the sheet of paper from reflected light images. Chain lines and laid lines can be detected from these type of images and its extraction can then offer future data for analysis at scale. These algorithms can be successfully applied to research in book history, bibliography and conservation. Our methods still require some manual fine tuning and parameter selection, however, future work will aim to fully automate the extraction pipeline to make the algorithms even more accessible to all scholars without any prior knowledge of coding or mathematical imaging.

\begin{ack}
TGG and ODR acknowledge the support of the Cambridge Humanities Research Grants Scheme and ODR acknowledges the ‘Thinking Paper Project’, Research and Collection Programme, University of Cambridge. TGG and CBS acknowledge the support of the Cantab Capital Institute for the Mathematics of Information and the European Union Horizon 2020 research and innovation programme under the Marie Skodowska-Curie grant agreement No. 777826 NoMADS. TGG additionally acknowledges the support of the EPSRC National Productivity and Investment Fund grant Nr. EP/S515334/1 reference 2089694. CBS acknowledges support from the Philip Leverhulme Prize, the Royal Society Wolfson Fellowship, the EPSRC advanced career fellowship EP/V029428/1, EPSRC grants EP/S026045/1 and EP/T003553/1, EP/N014588/1, EP/T017961/1, the Wellcome Trust 215733/Z/19/Z and 221633/Z/20/Z, and the Alan Turing Institute.
\end{ack}

\section*{Availability of data and materials}
The original images of paper manuscripts from The Parker Library, Cambridge, Corpus Christi College in Figures \ref{fig.abstract}, \ref{fig.chainline_extraction}, \ref{fig.chainline_result}, \ref{fig.laidline_result}a) and \ref{fig.laidline_anisotropic} are licensed under a \href{https://creativecommons.org/licenses/by-nc/4.0/}{Creative Commons Attribution-NonCommercial 4.0 License}. All images are openly available in their original form at \url{https://parker.stanford.edu/parker}. Images in Figures  \ref{fig.decomposition_example}, \ref{fig.laidline_extraction} and \ref{fig.laidline_result}b) are reproduced by kind permission of the Syndics of Cambridge University Library. Those images are available from the Cambridge University Library but restrictions apply to the availability of the images, which were used under license for the current paper, and so are not publicly available. The images are however available from the Cambridge University Library upon reasonable request.

The code for generating the resulting line extractions is openly available at \url{https://github.com/TamaraGrossmann/HiddenKnowledge}.

\bibliographystyle{abbrvurl}
\bibliography{bibliography}
\end{document}